\documentclass[preprint,12pt,3p,authoryear]{elsarticle}

\usepackage{lineno,hyperref}
\usepackage[usenames, dvipsnames]{xcolor}
\usepackage[title]{appendix}
\usepackage{soul}
\usepackage{amsmath}
\usepackage{float}
\usepackage{subcaption}
\usepackage{adjustbox}
\modulolinenumbers[5]
\usepackage{tabularx,booktabs}
\usepackage{todonotes}
\definecolor{sand}{RGB}{255,255,187}
\definecolor{shale}{RGB}{142,142,139}
\usepackage{amssymb}
\usepackage{ulem}

\journal{Journal of Petroleum Science and Engineering}

\begin{document}

\begin{frontmatter}

\title{Application of Machine Learning to accidents detection at directional drilling}

\author[label1]{Ekaterina Gurina\corref{cor1}\fnref{label4,label5,label7,label8}}
\cortext[cor1]{Corresponding author}
\ead{Ekaterina.Gurina@skoltech.ru}

\author[label1]{Nikita Klyuchnikov\fnref{label4,label5,label7,label8}}
\author[label1]{Alexey Zaytsev\fnref{label4,label5,label7}}
\author[label1]{Evgenya Romanenkova\fnref{label4,label5,label7}}
\author[label1]{Ksenia Antipova\fnref{label4,label5,label7}}
\author[label3] {Igor Simon\fnref{label6,label9}}
\author[label3]{Victor Makarov\fnref{label6,label9}}
\author[label1]{Dmitry Koroteev\fnref{label4,label9}}

\address[label1]{Skolkovo Institute of Science and Technology (Skoltech), 121205, Moscow, Russia}
\address[label3]{Gazprom Neft Science \& Technology Center, 19000, St. Petersburg, Russia}

\fntext[label4]{Research concepts}
\fntext[label5]{Implementation methods}
\fntext[label6]{Data collection}
\fntext[label7]{Data analysis and interpretation}
\fntext[label8]{Drafting the manuscript}
\fntext[label9]{Revising the manuscript}

\begin{abstract}
We present a data-driven algorithm and mathematical model for anomaly alarming at directional drilling. The algorithm is based on machine learning. It compares the real-time drilling telemetry with one corresponding to past accidents and analyses the level of similarity. The model performs a time-series comparison using aggregated statistics and Gradient Boosting classification. It is trained on historical data containing the drilling telemetry of $80$ wells drilled within $19$ oilfields. The model can detect an anomaly and identify its type by comparing the real-time measurements while drilling with the ones from the database of past accidents. Validation tests show that our algorithm identifies half of the anomalies with about $0.53$ false alarms per day on average. The model performance ensures sufficient time and cost savings as it enables partial prevention of the failures and accidents at the well construction. 

\end{abstract}


\end{frontmatter}

\section{Introduction}

Anomaly detection is the identification of rare objects, events, or observations that are significantly different from most data \citep{zimek2017outlier}. Regardless of the level of the well’s construction technology, anomaly situations inevitably happened during drilling. Anomalies may have both positive and negative influence on a system, depending on their interpretation and consequences. For example, a significant increase in the number of visits to a website might be considered as a positive anomaly, that results in website popularity growth. In the case of directional drilling, abnormal behavior rather leads to failures (emergencies which make any further work impossible or delay future activity), than to the improvement of the drilling process.

Accidents have a significant impact on the further operation of wells and usually lead to an increase in construction time and the cost of work. Drilling support engineers use mud logging to detect accidents while drilling. Due to the fact, that engineers support a large number of wells at the same time, they usually do not have time to monitor all the wells online, and drilling accident patterns are considered after the occurrence of the accident. Thus, the creation of a system that signals about drilling accident will help engineers more efficiently support the drilling process. Early detection of failures can significantly reduce the nonproductive time of the well associated with the elimination of accidents’ consequences and costs for additional materials and technical resources. Most of the oil and gas companies also create knowledge base systems, in which information about failures is collected and carefully studied in order to use accumulated experience for further failures detection by comparison current drilling conditions and previous cases. Such an approach is called analogues search and was successfully used for time-series forecast \citep{analogues_search_meteo, analogues_search_2}. Thus, the presence of extra support during drilling operations as an expert system, that includes accumulated accident detection experience, is an effective method for making the right real-time decisions. Such a system will allow to avoid additional expenses during drilling operations, and reduce the high workload level of the drilling engineers. Therefore, the development of the methods, that can help to detect failures during the real-time drilling operations is essential for the oil and gas industry. 

This paper is aimed at the analysis of the analogues search approach for application to drilling accidents detection. In more details, the key objectives of the article are the development of the model, that will be able to distinguish similar and non-similar drilling situations, the analysis of model’s applicability limits, and quality.

The main contribution of this paper is an anomaly detection approach for directional drilling operations, called the analogues search model. It is designed for ranking the accidents from the knowledge base according to their relevance to the current situation in the drilling process, in order to find analogues and prevent anomalous behaviour. The solution is based on the compassion of mud logging data with the classification model built on Gradient Boosting of decision trees. 

\subsection{State-of-the-art}
\label{sec:2}

Anomaly detection is an important issue that has been investigated in various research areas: there are some examples of anomaly detection in Information Technology systems \citep{IT_article}, medicine \citep{Medicine_article} and industry \citep{Safety_article}. In oil and gas industry anomaly detection is widely spread: in downstream it is used for controlling pumping and pressure in different systems, drilling process \citep{Oil_and_gas}, lithology classification \citep{litho_1,litho_2}; in upstream, for example, engineers usually use it for detection sensors faults in a refinery \citep{refinery} and pipelines \citep{Pipelines}.  

Solving the problem of unusual behavior detection in drilling by analogues search, it is necessary to consider not only previous studies on time-series comparison and general algorithms of anomalies detection but also methods and approaches for accidents detection during drilling, since they often happen as a result of anomalies.

\subsubsection{Methods for time-series comparison}

Considering the problem of analogues search, it is necessary to compare different time-series. Several authors  \citep{Similarity_time_series, Similarity_time_series_2, Similarity_time_series_3} suggest to measure the similarity between two time-series by different metrics, for example, general Euclidean distance, Fourier coefficients, the Time-wrapping (TW) distance, and its modifications.

After the introduction of any distance, the whole database of time-series can be split into several groups with different clustering techniques, for example, K-means algorithm \citep{kanungo2002efficient}, mean-shift clustering \citep{cheng1995mean}, agglomerative hierarchical clustering \citep{day1984efficient}. Authors highlighted, that general Euclidean distance and Fourier coefficients showed themselves inefficient for time series with different length,  while the cost for Time-wrapping distance computation for $m$-dimensional time-series might be significant. 
For our case, mud logging patterns for different accidents and different oilfields are too diverse to apply such metrics effectively. 
Thus, clustering of raw time-series seems incompetent for analogues search problem, which makes us move to supervised approach for similarity learning based on statistical features extracted from time-series.

\subsubsection{Methods for anomaly detection}
Most of the general methods for anomaly detection were described previously, for example, in papers \cite{sliding_window, ML_overview, IT_article}. Authors distinguished several groups, based on statistical methods, machine learning, and unsupervised approaches. Due to the high variability of general methods and a large number of a review papers on them, we will not focus on description and will show only a few examples, which are relevant to the analogues search problem. 

In paper \cite{sliding_window} authors describe the approach, based on sliding window technique, in which some parts of time series with width $w$ is converted into a single target value $y_i$ by some particular classifier. By this principle, the sequences of signals were classified for the whole time-series signal as an anomaly or non-anomaly target value. 
The main advantage of this method is the possibility of applying different existing classification methods. 
For anomaly detection in drilling, such an approach allows us to convert the unsupervised approach into supervised one, but do not involve physics of the drilling, which is significant for drilling accidents detection problem. 

Nowadays, there are a lot of cases of neural networks \citep{Neural_basic} applications for anomaly detection \citep{neural_net_2, neural_net_3}. For example, authors in \cite{DL} used a neural network to hierarchically learn features from the sensor measurements of exhaust gas temperatures and used them as the input to a neural network classifier for performing combustor anomaly detection. As a training set, the authors used $13 791$ samples before the accident. In our case, this approach may be inefficient due to the small size of the training sample and inability automatically handle missing values, which usually occur in mud logging data. 

In article \cite{IT_article} authors highlight such approaches as a deviation of normal behavior and statistical methods. For example, \cite{normal_behaviour} collected the stable database of activities not leading to intrusions and then used it to analyse the current behavior of the system by its comparison with database modes by different statistics.  Comparing this approach with the problem of drilling accidents detection, it can be noticed that this approach is almost impossible to use, because, unlike the user system, each well and field is unique. Usually, a similar slight deviation of normal drilling regime in one well can lead to serious accidents on the other.

\subsubsection{Physics-based methods for drilling accident detection}

Physics-based methods for detection accidents are primarily based on the monitoring and analysis of the key indicators of the drilling system. For example, 
\cite{physical_indicators} describes physical indicators and their changes, leading to failures. One of the main indicators of fluid shows while drilling are an increase in the volume of the drilling mud in the receiving tanks, a reduction of standpipe pressure, an increase in the effluent flow rate with a constant flow of pumps, an unexpected increase in the mechanical penetration rate (due to a decrease in the density of the drilling mud, and, consequently, the pressure in the well). In case of wash-outs, the main evidence of the accident is an increase of a drill string weight, friction reduces, reduction versus the calculated volume of fluid full up when lifting the pipe string, movement of the drilling mud along the ditch system with the circulation stopped.

One more example of a physical-based method for failure detection is vibration, namely modeling the movement of the drill string and its components, which is represented in paper \cite{vibration}. Early models of drill string dynamics have been developed primarily as an aid to drilling engineers and rig designers, to help them understand wells behavior and provide recommendations for improving the drilling operations. Currently, models are being used and investigated based on three parallel but different vibration modes, those help engineers detect anomaly by high vibration values.

Due to the inability to track all the indicators above, such physical-based methods are not suitable for solving our problem.

In addition to the methods shown above, there are a different anomaly and failure patterns in mud logging plots. For example, a high number of drags and slack off are used as signs of a possible pipe stuck. Column drags usually occur while the column was lifting with an increase a hook load over its weight of pipes; the slack off of the tool results in a significant reduction in the load on the hook. One of the evidence of columns stuck is also stopping of the column movement. In the case of wash-outs, a decrease in pressure at a constant flow rate might be observed (\citep{ Washouts}). The main failure pattern characterising the mud loss is a decrease in the volume in the tanks. Breakdown of the tools is marked by a reduction in pressure at a constant flow rate simultaneously with a sharp drop in weight. So, particular patterns for each type of accidents on mud logs might be used as the first signs by which the model can determine the presence of failure.

\section{Methodology}

\subsection{Data overview}
To solve the problem of failures detection by analogues search approach, a database with different types of accidents and their mud log data was collected.  

Most of the failures happened in North and West Siberia oilfields and were composed of accident lessons that contain the information about these events: the exact date-time or depth at which the failure occurred. Such criterion was chosen in order to match the mud log data with the accident from the database and get a part of it that includes the failure. 
Each lesson included in the database also contained information about its accident type (stucks, wash-outs, breaks of drill pipe, mud loss, shale collars, gas, and water shows) and drilling operation at the moment of failure (tripping in, tripping out, drilling, cleaning, reaming). 
Such groups of accidents and drilling operations were chosen by the number of available cases and a possibility to be distinguished visually on mud logs.

In total, the database contains $94$ lessons from $80$ different wells and $19$ oilfields. 
The summary of the size of different considered groups of accidents and related drilling operations is provided in Table \ref{tab:summary_database}.

\begin{table}[ht]
 \centering
    \begin{tabular}{ccccccc}
    \hline
     & Triping in & Tripping out & Drilling & Cleaning & Reaming & Total \\ \hline
    Stuck & 18 & 11 & 10 & 0 & 1 & 40 \\ 
    Wash-outs & 1 & 1 & 10 & 1 & 0 & 13 \\ 
    Breaks of drilling & 1 & 2 & 4 & 6 & 0 & 13 \\ 
    Mud loss & 2 & 2 & 6 & 0 & 1 & 11 \\
    Shale collars & 0 & 0 & 9 & 0 & 0 & 9 \\ 
    Fluid shows & 0 & 3 & 5 & 0 & 0 & 8 \\ \hline
    Total & 22 & 19 & 44 & 7 & 2 & 94 \\
    \end{tabular}
    \caption{Breakdown of included accidents by type of accident and phase of drilling: in some cells we have almost no example for training}
    \label{tab:summary_database}
\end{table}

The considered measurements while drilling (MWD) data included depth of the drill bit, torque on the rotor, weight on the hook, input pressure, rotation speed, a volume of input flow, a depth of the bottom hole, gas content, and weight on bit.

\subsection{Design of the analogues search model}

In section \ref{sec:2}, we discussed existing approaches for time-series comparison, anomaly, and failure detection.
It was concluded, that for the problem of a drilling accident detection, it is necessary to use a supervised machine learning approach. The algorithm should take into account the particular mud logs pattern for different accident groups and be able to work with a small training set and corrupted or missing signal values.

We decided to solve the analogues search problem based on two-class classification of MWD pairs: for a specific well part, we need to understand whether something similar is present in the database by comparing features from MWD data of this part with those of entries in the database. Thus, there are two classes that determine whether two parts are similar or not.

For the current approach, we decided to build a classification model based on Gradient boosting of decision trees, because they are relatively undemanding in terms of sample size and data quality, can work with missing data, and learn quickly with a large number of features, what was shown in papers \citep{Grad_boost, Grad_boost_2}. 

\begin{figure}[ht]
    \centering
    \includegraphics[width = 0.9\linewidth]{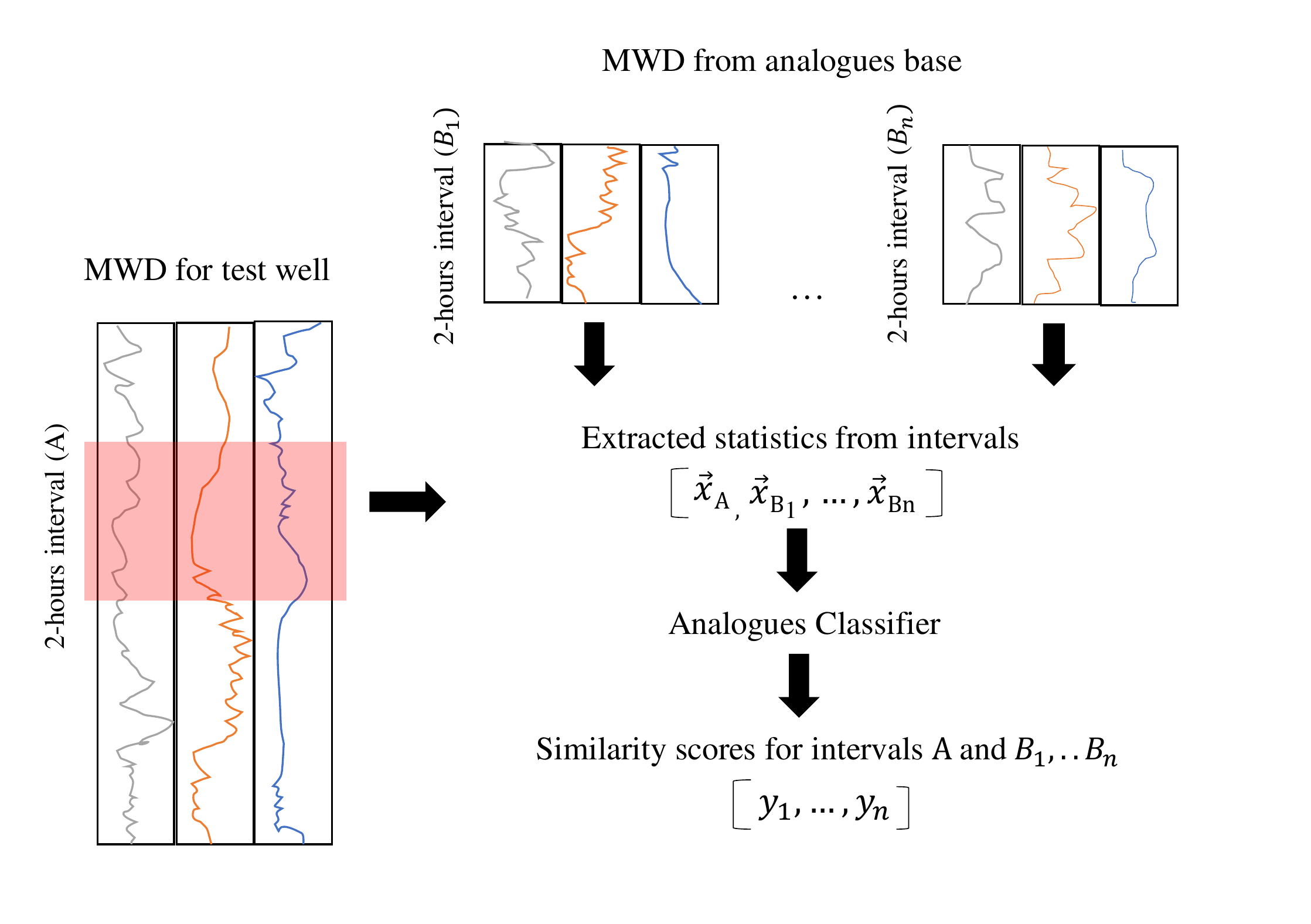}
    \caption{General scheme of analogues search model. Using $2$ hours parts of MWD signals, different aggregated statistics were calculated. These features are inputs for gradient boosting classifier, that provides similarity scores for a pair of input signals.}
    \label{fig:general_sheme}
\end{figure}

The general principle of analogues search model is shown in Figure \ref{fig:general_sheme}. In order to take into account different patterns, for real-time signal and lessons from the database values of mean, variance, slope angle, absolute deviations, and relative coefficients of MWD time-series were calculated with different window sizes and were used as \textit{input features} for Gradient boosting classification model. To assign targets, we assumed that pairs of intervals were similar if their accident types and drilling operations coincided. Henceforth we will refer to them as to \textit{ground truth}

To test the analogues model, several experiments were conducted. In order to validate our model, the standard quality metrics for the binary classification problem using leave-one-out cross-validation control were calculated. We also carried out a clustering analysis based on similarity values from the model to validate the aggregated statistics approach and evaluate the consistency of the similarity learning. The similarity distributions between MWD data with accidents and random MWD parts of wells without abnormal behavior were analysed, in order to assess the model ability to distinguish regular drilling regime and accidents. In addition, we provided a sensitivity analysis with respect to various kinds of noise in MWD data.

\section{Results and discussions}
\subsection{Quality of the analogues search model}
\label{sec:quality}

For the analogues search model, the cross-validation was carried out as follows:
\begin{enumerate}
    \item For each of $k$ iterations of cross-validation, random indexes of accidents from the database were generated for training and testing sets. The set of wells for accidents were different for training and test part of the split.
    \item The model was trained based on lessons, which indices were chosen as training ones. 
    \item Similarity values among entries in the training and test set were calculated. The model finds the analogue and, consequently, detects failure, if the similarity value is bigger than the selected threshold ($ s = 0.7$).
    \item The predicted values were compared with ground truth labels.
\end{enumerate}

The results of cross-validation for analogues search model are in Table~ \ref{tab:confusion_matrix}.
Using the current model, it is possible to identify almost all wells with abnormal drilling regime with low false alarm rate. So, it can be concluded that the model distinguishes different pairs quite well and identifies most of the similar ones.

\begin{table}[H]
    \centering
    \begin{tabular}{c|cc}
     & Predicted = 1 & Predicted = 0 \\
    \hline
    True = 1     & 5792 & 294 \\
    True = 0     & 223  & 345 \\
    \end{tabular}
    \caption{Confusion matrix for threshold $s = 0.7$}
    \label{tab:confusion_matrix}
\end{table}

\begin{figure}[H]
\begin{subfigure}[!ht]{0.47\linewidth}
\center{\includegraphics[width=1\textwidth]{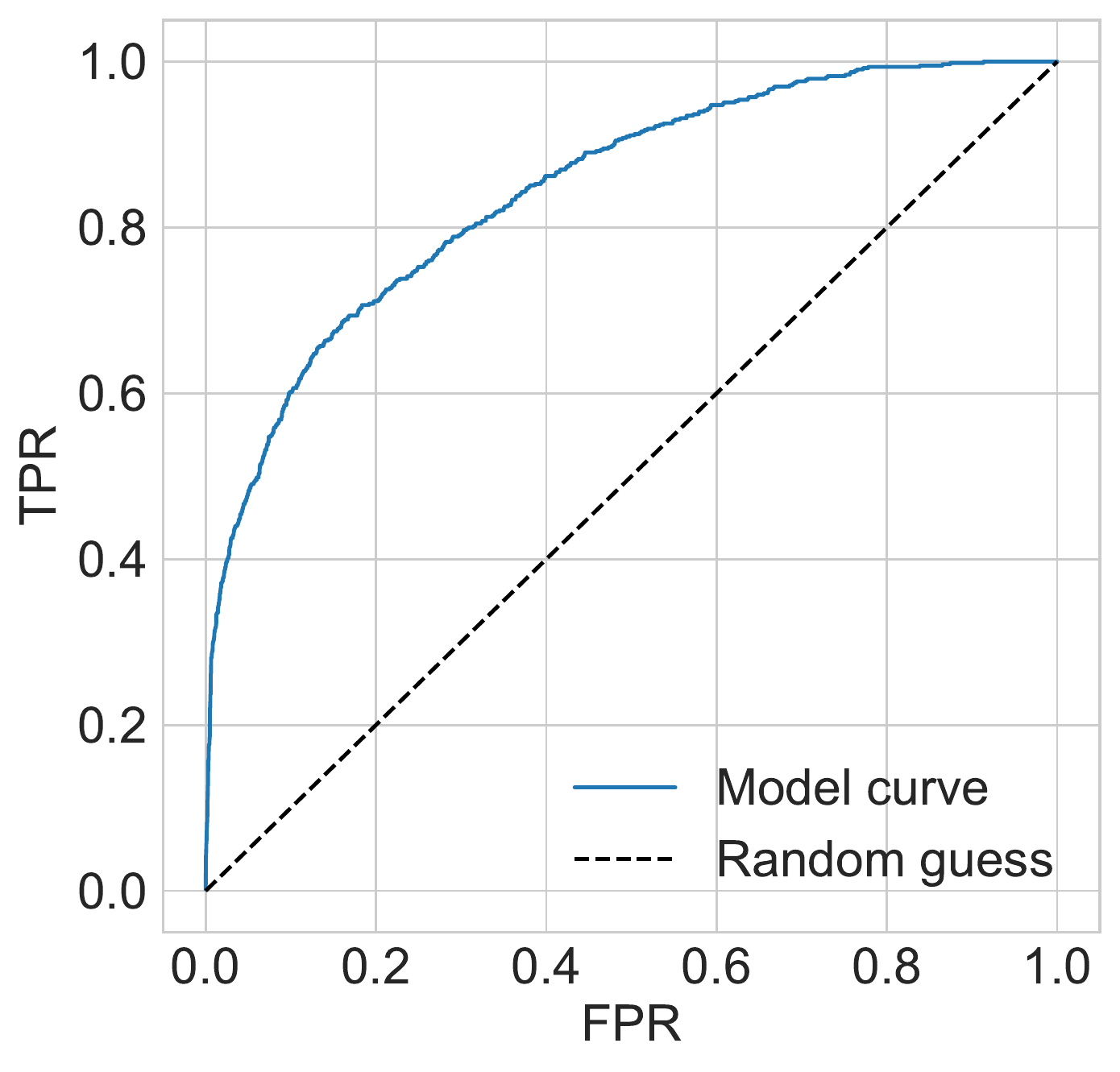}} 
\caption{ROC curve}
\end{subfigure}
\hfill
\begin{subfigure}[!ht]{0.47\linewidth}
\center{\includegraphics[width= 1\textwidth]{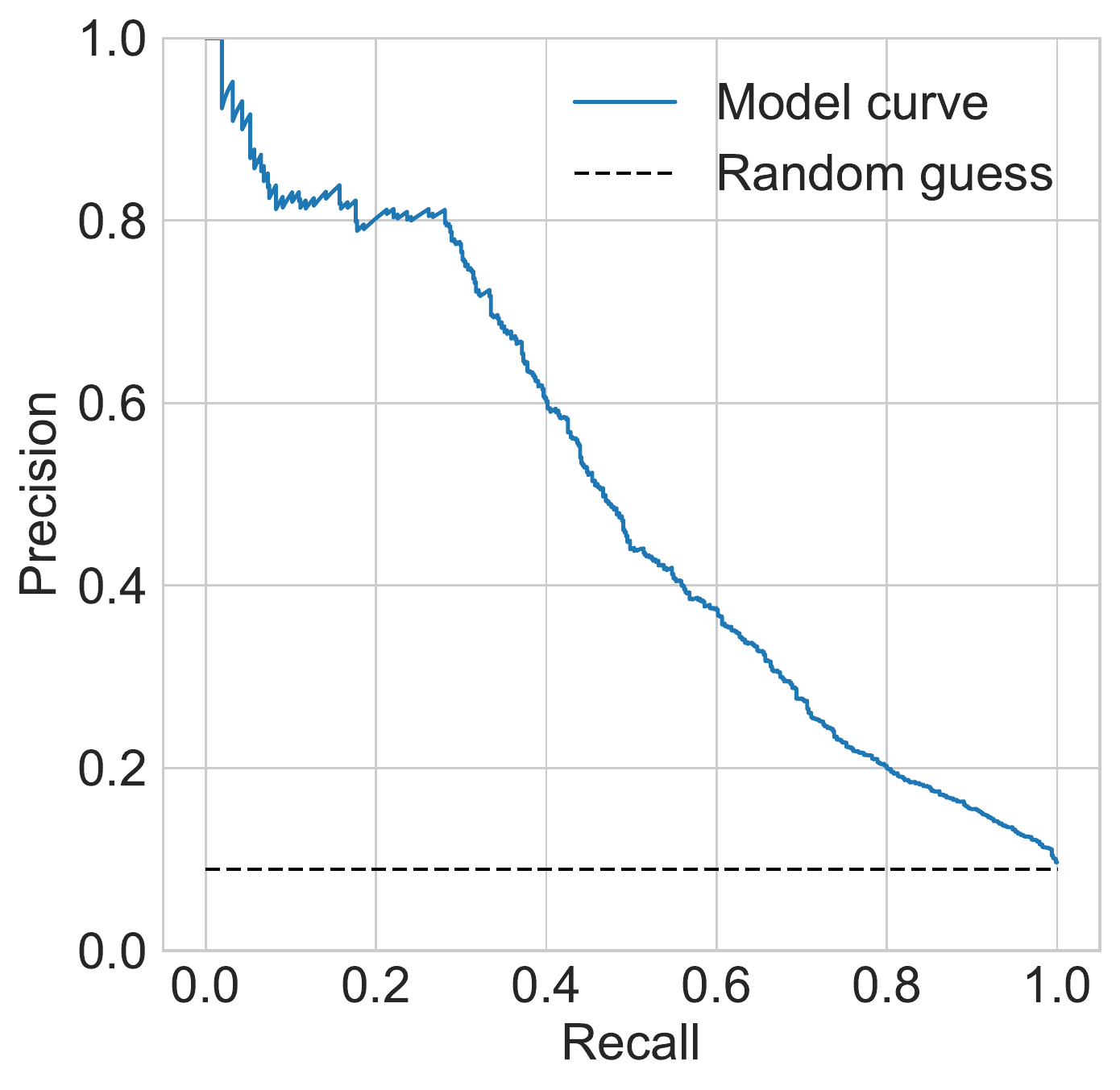}} 
\caption{Precision Recall curve}
\end{subfigure}
\caption{Quality metrics for analogues search model. ROC AUC is $0.908$, thus the model is significantly better than a random guess with ROC AUC $0.5$. The area under PR curve is $0.6086$, which is significantly better than the area under curve for a random guess approach $0.1$.}
\label{fig:quality_metrics}
\end{figure}

In order to obtain the model quality, two common metrics for classification problems were used: area under the receiver operating characteristic curve (ROC AUC) and area under the Precision-recall curve (PR AUC), which are described in details in Appendix \ref{Appendix_2}. The receiver operating characteristic (ROC) curve is presented in Figure \ref{fig:quality_metrics}. The area under ROC curve is $0.908$, and significantly higher, than the area under the random guess classifier ROC curve  $ 0.5 $. Since it is an unbalanced classification problem, a more suitable measure of model quality will be a Precision-Recall curve, which is shown in Figure \ref{fig:quality_metrics}. The area under the Precision-Recall curve is $ 0.6086$, which also indicates adequate model quality.

\subsection{Hold-out validation and threshold selection by analysis of confusion matrix}

In this section, we used the analogues search model differently: it was applied to hold-out wells in order to understand how it works “in the wild.” The analogues model was run on MWD signals from $30$ hold-out cases, which included both normal and anomaly drilling modes. 

Next, the threshold to balance the number of correct ($\mathcal{TP}$) and false ($\mathcal{FP}$) alarms was selected. After that, based on the true accident time for each well, the true and false model’s alarms rate were calculated as follows. We assumed, that accident was correctly detected, and for this accident $\mathcal{TP} = 1$, if the similarity value was more than the chosen threshold, the model alarm was in the $4$-hour interval before and $2$ hours after the true accident ($\mathcal{TP}$ interval), and the most common accident type for the top-5 analogues matched with true one.

In case of the false alarm, it was supposed that $\mathcal{FP} = 1$ for this interval if the model alarm was out of $\mathcal{TP}$ interval and there were no other alarms during the last hour. So if two or more alarms happened within 1 hour, it was counted as one false alarm. Here we also assumed, that predicted accident type was the most common one within top-5 analogues types; otherwise, it is supposed that $\mathcal{FP} = 1$. Obtained results for the threshold $0.7$ are presented in Appendix~\ref{Appendix_1}. 

\begin{figure}[H]
    \centering
    \includegraphics[width =  \linewidth]{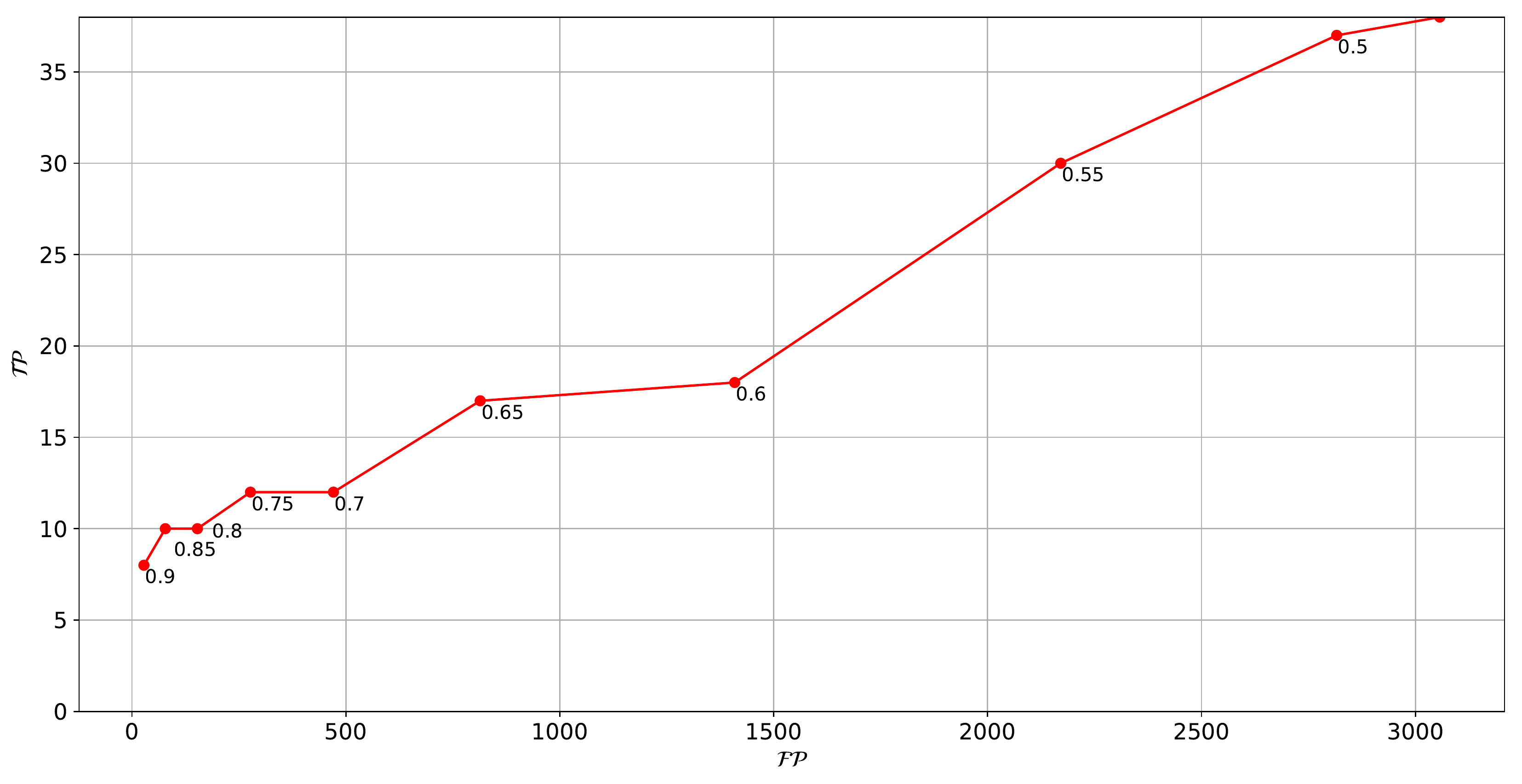}
    \caption{Total number of model correct (True Positive, TP) and false (False Positive, FP) alarms for different thresholds. Numbers at the curve are thresholds. The threshold $0.7$ was used on final model.}
    \label{fig:trade_off}
\end{figure}

To select the threshold, the total number of $\mathcal{TP}$ and $\mathcal{FP}$ for different threshold values was counted (Figure \ref{fig:trade_off}). For the threshold value $0.7$, the total number of model false alarms is less than $16$ alarms per well, while the number of correct alarms is still high. 

Let us consider one example of analogues search model results for a hold-out well. In this case, the model ran on a well that contained a wash-out drilling accident. In this case, the model found an analogue with the same type of operation (drilling) as in the hold-out MWD measurements. 
It can be seen based on the general similarity of trends in such parameters as hook position, depth of drill bit, and bottom hole depth. In cases, when similarity values exceeded the selected threshold, the model assumed, that current two hours in the past are similar to the analogue measurements. For both cases, we observe a decrease in pressure at a constant flow rate, which indicates the wash-out of the drill pipe. A careful examination of this case proves that the model correctly detected a wash-out accident and found an analogue.  
Similarity values and MWD signals for the "current" measurements and the analogue are presented in Appendix~\ref{Appendix_b}.

\subsection{Clustering analysis}
\label{sec:clustering}

The dendrograms clustering analysis \citep{Dendrogram} was also used to assess the consistency of the similarity learning. First of all, we represented it via adjacency matrix clusters based on the ground truth distribution of similarity. As mentioned earlier, two lessons are similar if their accident types and drilling operations are equal. Next, to compare initial distribution, similarity values that were used as an input parameter for constructing dendrograms were calculated in different ways:
\begin{itemize}
    \item {\it Unsupervised comparison:} similarity values for lessons from the database were calculated only by the weighed $l_1$ norm among all MWD parameters, excluding the depth of the bottom hole and drill bit.
    \item {\it Using Gradient boosting technique:} dendrograms used similarities, that were calculated for lessons from the training set, and resulted from the Gradient boosting model with aggregated statistics. This is an optimistic estimate of the quality of the similarity evaluations.
    \item {\it Cross-validation:} calculations were made with the model described in the previous step (using Gradient boosting technique) and cross-validation, which allows us to see how well the model generalise to new cases of accidents. This is a more realistic estimate of the quality of the similarity evaluations.
\end{itemize}

\begin{figure}[!ht]
\begin{subfigure}[!ht]{0.47\linewidth}
\center{\includegraphics[width = 0.9\linewidth]{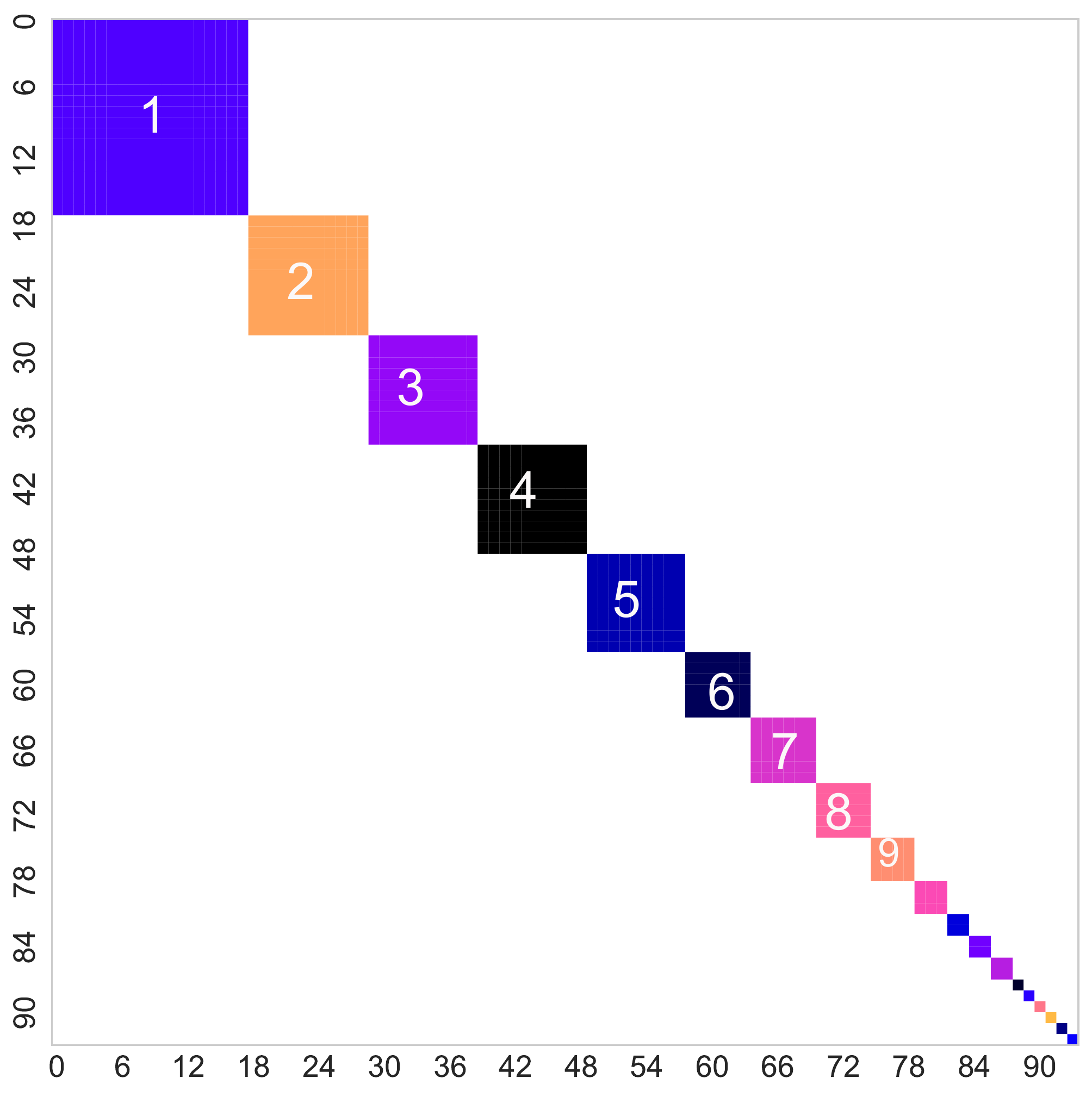}} a) \\
\end{subfigure}
\hfill
\begin{subfigure}[!ht]{0.47\linewidth}
\center{\includegraphics[width = 0.9\linewidth]{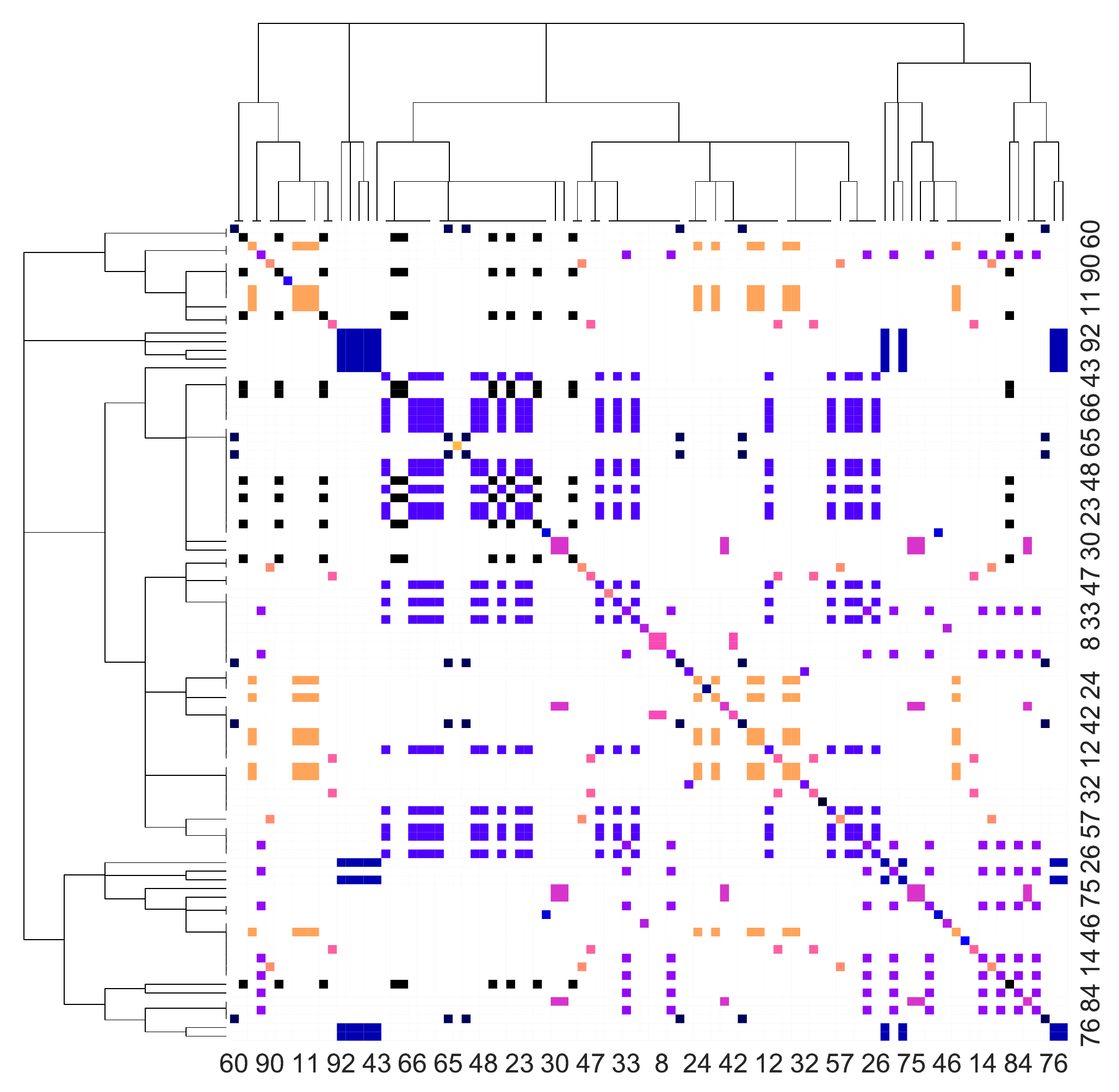}} b) \\
\end{subfigure}
\vfill
\begin{subfigure}[!ht]{0.47\linewidth}
\center{\includegraphics[width = 0.9\linewidth]{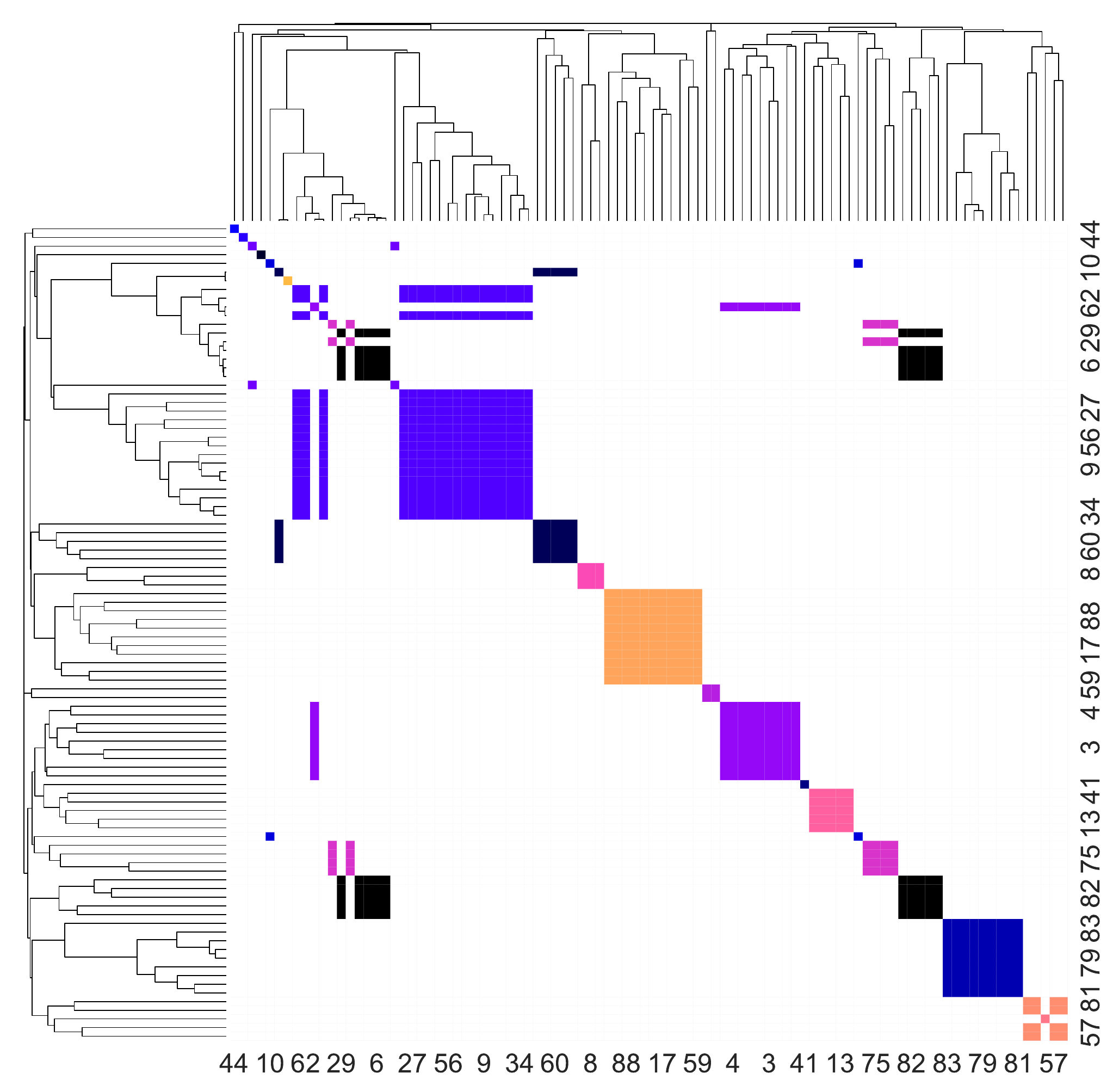}} c) \\
\end{subfigure}
\hfill
\begin{subfigure}[!ht]{0.47\linewidth}
\center{\includegraphics[width = 0.9\linewidth]{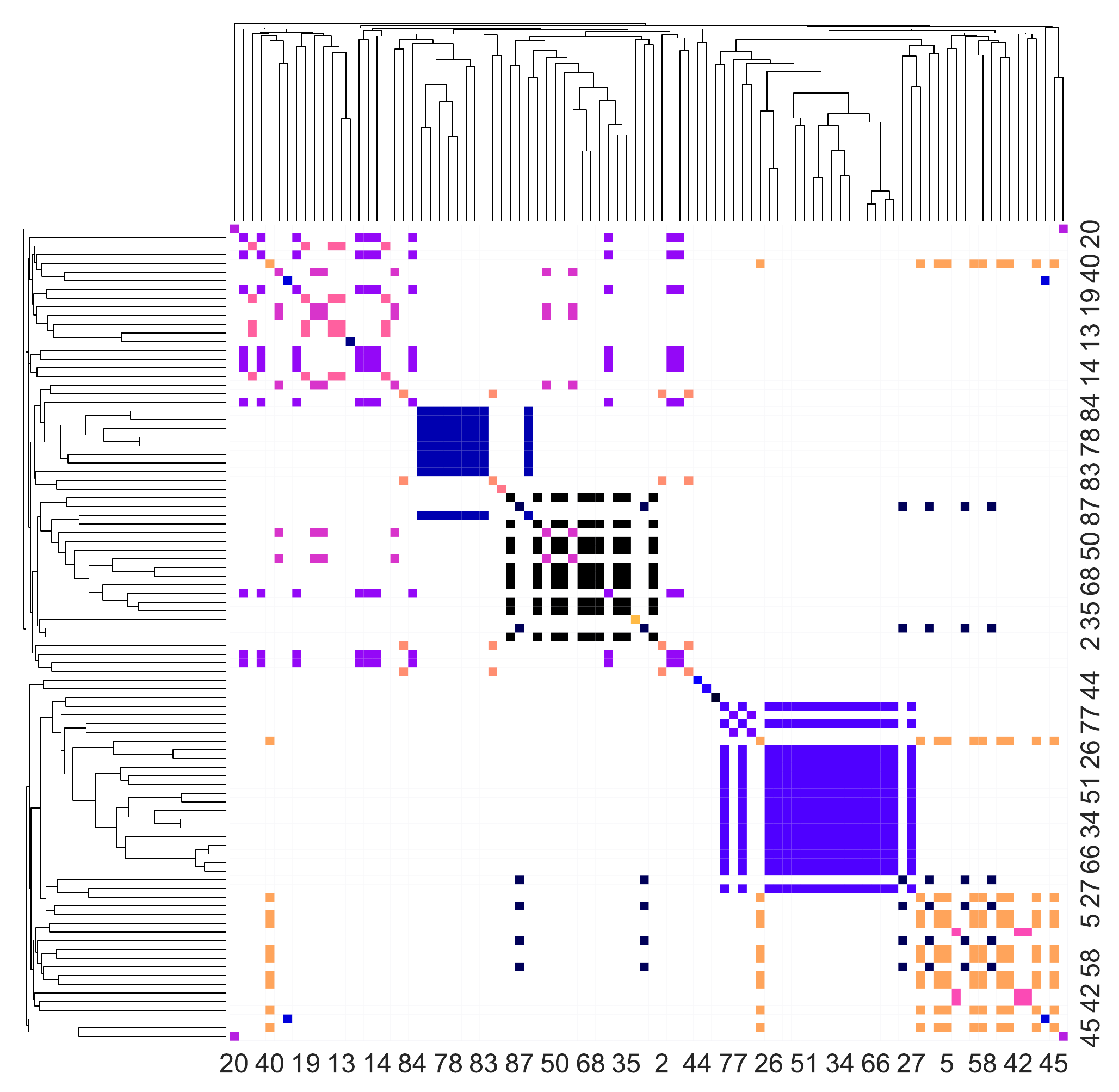}} d) \\
\end{subfigure}
\caption{Clustering analysis: a) Initial clusters distribution, b)
Dendrogram, based on simple comparison of MWD data, c) clusters, obtained from aggregated statistics and gradient boosting technique, d) dendrogram, obtained from cross-validation. Clustering analysis represented by dendrograms, which illustrates how each cluster is composed by drawing a link between clusters: the top of the link indicates a cluster merge, while the two legs indicate which clusters were merged. In our case, each true group of accidents correspond to one colour, and is formed by the same type of drilling accidents, which have the same operation type. The presence of colour at the intersection of row $i$ and column $j$ means that two cases,$i$ and $j$ respectively, of drilling accidents from the database belong to the same true accident group. Since the main aim of this experiment is to see if similar anomalies can be grouped into clusters using different approaches, instead of the global order of clusters it is important to observe local proximity of elements. Using aggregated statistics as an input for Gradient boosting classifier, the clusters distribution is more similar to the initial one, and it is possible to obtain more separated clusters, than by using only raw signals.}
\label{fig:clustering}
\end{figure}

The results of the conducted test are shown in Figure~\ref{fig:clustering}. Usage of aggregated statistics with Gradient boosting model gives us a better cluster distribution than approaches based on an unsupervised comparison of MWD data. Results, obtained from cross-validation, shows us clearly selected clusters corresponding to different types of accidents and drilling operations. For some types of accident, the training set is quite complete, that can be seen by the trees above and to the left of the plot, and has enough cases (clusters number 1,3,4,5). For others (clusters number 2,6-9), there is a greater distance for objects within the cluster. In our opinion, the reason for this might be the lack of examples of accidents in these groups. Consequently, for the correct determination of such groups of failures, the inclusion of a larger amount of data is required.

\subsection{Robustness of the analogues search model}
\label{sec:robustness}

The analogues search model should meet the following two requirements. If an example from the training sample was submitted, the model should recognise it and provide it as the analogue with high similarity. Moreover, after reasonable distortion of such an example, the model should still recognise it.

\begin{figure}[!ht]
    \centering
    \includegraphics[width =\linewidth, scale = 0.9]{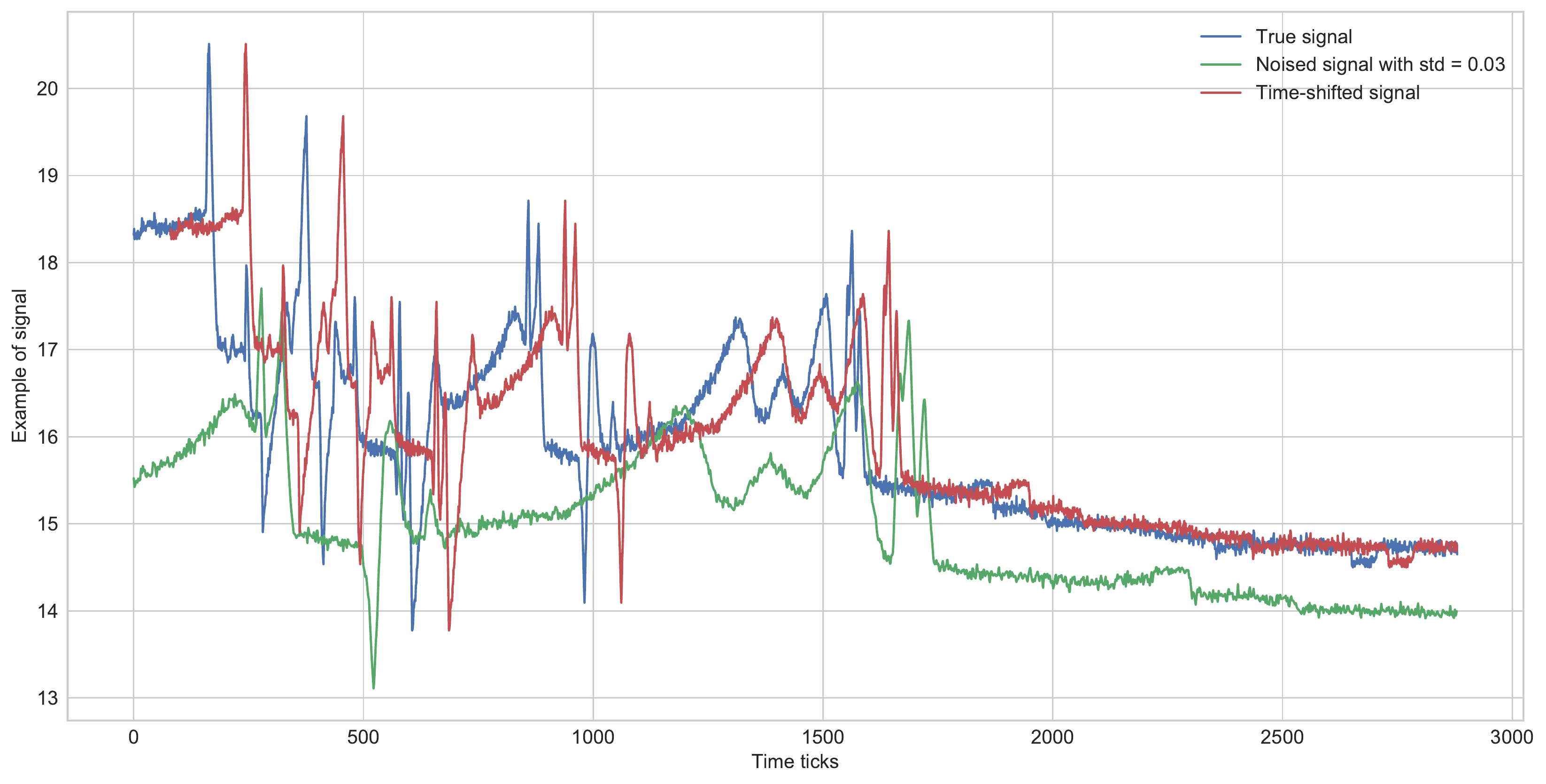}
    \caption{An example of original and distorted time series values}
    \label{fig:noised_curve}
\end{figure}

While testing the first property is straightforward, to test the second one-two types of transformations were applied to the original time series: slight smoothing, distortion, and shift of data on given number of time ticks (1 tick = 10 seconds). An example of the original and distorted time series for different values of the standard deviation is given in Fig \ref{fig:noised_curve}.

To understand how well the model distinguishes MWD parts of wells with normal behavior and ones with accidents, the model was trained on MWD parts, corresponding to the lessons from the current database and tested for normal and distorted parts. 

The distribution of obtained similarity values was presented as box-plots for different testing sets: random parts without accidents, intervals with accidents, time-shifted duplicates of the intervals with accidents and copies of the intervals with accidents with varying levels of shift and noise. The distortion of original time-series was done by the multiplication of a smooth curve with average mean $1$ and a given standard deviation. 

The numerical characteristic $R$ was also calculated: the difference between the $90\% $ quantile of random parts set and the $ 10\%$ quantile of data that we would like to highlight.
Valid values are bigger than $0$; good ones are more than $0.2$. Standard deviations for $R$ were calculated using the bootstrap technique \citep{efron1986bootstrap}, the calculation used $100$ samples.

\begin{figure}[!ht]
    \centering
    \includegraphics[width =\linewidth, scale = 0.9]{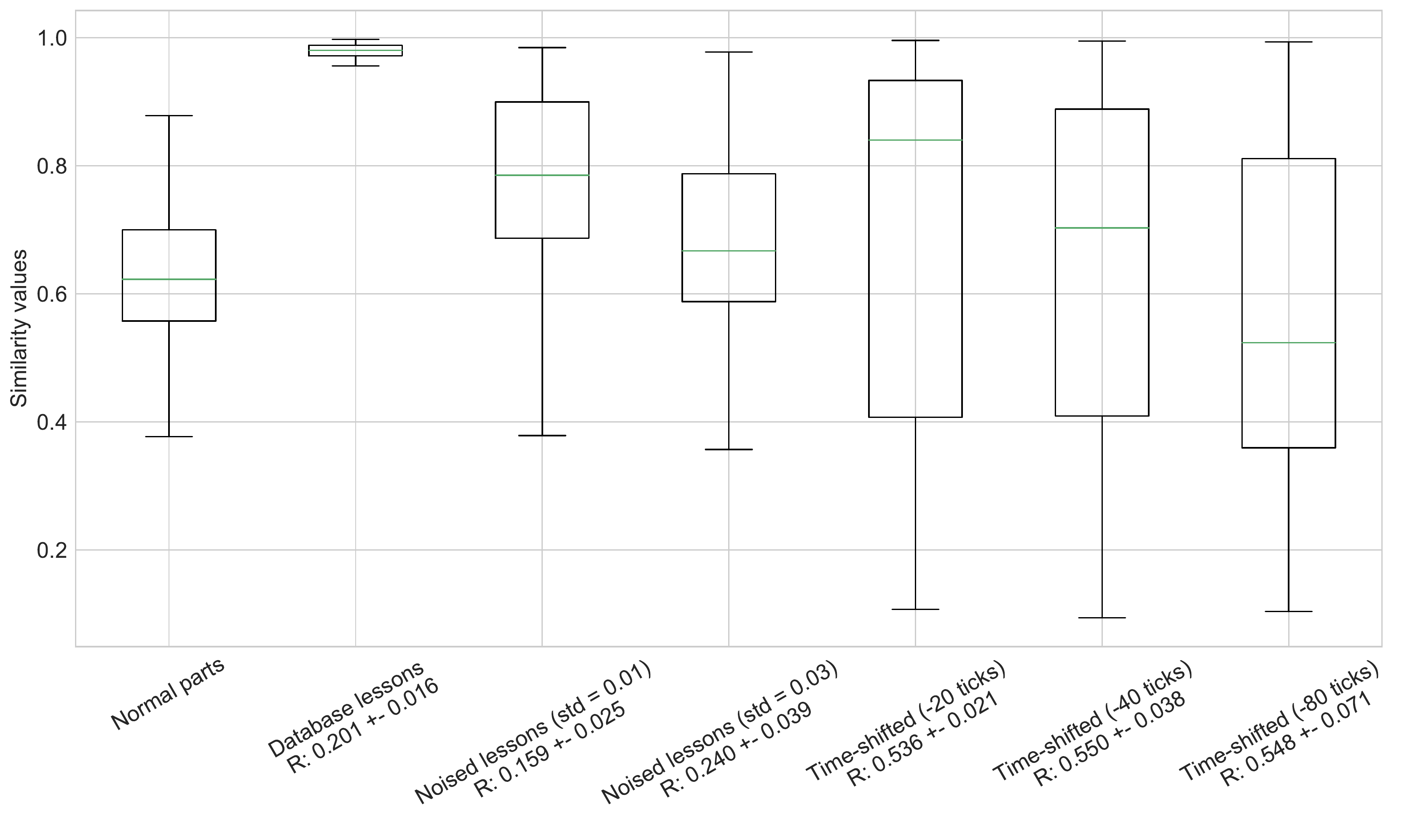}
    \caption{Box-plots for different intervals. Such figure gives us an idea of how much MWD data can be distorted, so that the model can still recognise them}
    \label{fig:box_plots}
\end{figure}

The box-plots for cases, mentioned above, and values of $R$ coefficient, which characterises the difference in the similarity values for two different sets of intervals are in Figure \ref{fig:box_plots}.

The analogues search model shows high similarity values for noised lessons with standard deviations as high as $0.01$ and time-shifted lessons as high as $20$ ticks. So, in these cases, the model finds similar sections from the training set.

At the same time, the similarity values for normal parts are low, which shows the model ability to distinguish normal drilling mode from the accidents-related drilling mode. It also can be seen that model can separate random MWD parts from the data, corresponding to the lessons from the database, for shifting up to $400$ seconds and for noise with a standard deviation of up to $0.03$. 

\subsection{Discussion}

When a drilling engineer encounters an accident, he or she can identify the type of the accident and rely on actions taken during similar cases in the past. Moreover, from now, it is rather evident, that it is possible to correlate the real-time state of the drilling process with the past events from the historical database with machine learning (see sections \ref{sec:quality}, \ref{sec:clustering} and \ref{sec:robustness}).

The quality of the machine learning model is acceptable for use during drilling as even a single successful application of the developed model can save time and money. However, there is still room for improvement. In particular, the model quality for identification of types of accidents underrepresented in the database is sometimes low. Thus, to maximize quality, a larger database is required.

In the separate section, we address the issue of model reliability and generalization ability. Conducted tests suggest a limit of applicability of the model, as after significant distortion of the initial signal the model no longer works correctly. However, the obtained results correspond to the general machine learning theory of extrapolation and generalization properties of the data-driven models.

\section{Conclusions}

We developed a real-time analogues search model that detects anomalies and finds analogues in a database of historical data.

The anomaly detection is based on the smart comparison of the real-time MWD data and the MWD data from the historical database followed by the ranking of the lessons from the database by their similarity to the real-time state. The comparison and the ranking utilize Gradient boosting classification model.

Conducted analysis of analogues search model showed that obtained quality metrics, such as ROC AUC ($0.908$) and PR AUC ($0.6086$), are significantly higher than the same metrics for the random guess classifier (ROC AUC: $0.5$, PR AUC: $0.1$). Obtained metrics suggest that the model is of reasonable quality and can distinguish pairs of similar and non-similar cases well. 

Clustering analysis showed that the use of basic MWD signals is not sufficient for the selection of analogues, and in general for the analogues search model. We have discovered that the introduction of aggregated statistics as input for Gradient boosting classifier allows finding a sufficiently larger number of analogues of real-time signals.

According to the robustness analysis, the model identifies lessons from the training sample, if such lessons are also in a testing sample. The analogue search quality remains high even after reasonable distortion of examples from the training sample by adding noise and shifts to the initial signal. These experiments helped to identify the limits of applicability of the model, ensuring a good understanding of what level of MWD signal distortion is still acceptable for an accurate analogues identification.

We plan to expand the functionality of the model and conduct research on whether it is possible to predict the failures beforehand.

\bibliography{mybibfile}

\begin{appendices}
\section{}
\label{Appendix_2}

To define  area under the ROC curve (ROC AUC) and area under the Precision-recall curve (PR AUC) metrics it is necessary to introduce additional notations. Suppose there are the true labels of the classes $ \{y_i \}_{i = 1}^{n}$ and the predicted labels of the classes $ \{\hat{y} _i\}_{i = 1}^{n}$ for a test sample.
Then the following 4 indicators sufficiently reflect the quality of classification: number of true classified objects for the first class (True Positive), the first class objects, that were classified as objects from the second class (False Negative), the second class objects, that were classified as objects from the first class (False Positive), true classified objects for the second class (True Negative). 

\begin{eqnarray}
    True\ Positive(TP) =\sum_{i = 1}^n [y_i = 1] [\hat{y}_i = 1], \\
    False\ Positive(FP) =\sum_{i = 1}^n [y_i = 0] [\hat{y}_i = 1] \\
    False\ Negative(FN) =\sum_{i = 1}^n [y_i = 1] [\hat{y}_i = 0],\\
    True\ Negative(TN) =\sum_{i = 1}^n [y_i = 0] [\hat{y}_i = 0] 
\end{eqnarray}

Based on such indicators, it is possible to introduce derived metrics: True Positive Rate (TPR) and False Positive Rate (FPR). 

\begin{eqnarray}
TPR =  \frac{TP}{TP + FN}, \quad FPR = \frac{FP}{FP + TN}, \quad Precision = \frac{TP}{TP + FP},  \quad Recall = TPR 
\end{eqnarray}

By setting the threshold, fixed $TPR$ and $FPR$ for the test sample were obtained. Varying the threshold, we get a ROC curve that starts at $(0, 0) $ and ends at $(1, 1)$. Similarly, the PR curve defined by the (Precision, Recall) values for a set of thresholds. ROC AUC, PR AUC are the areas under the curves, respectively, ROC and PR. The higher the values of these metrics, the better the quality of the classifier \citep{metrics_2}.

\section{}
\label{Appendix_b}
\begin{figure}[H]
    \centering
    \includegraphics[width =\linewidth
    ]{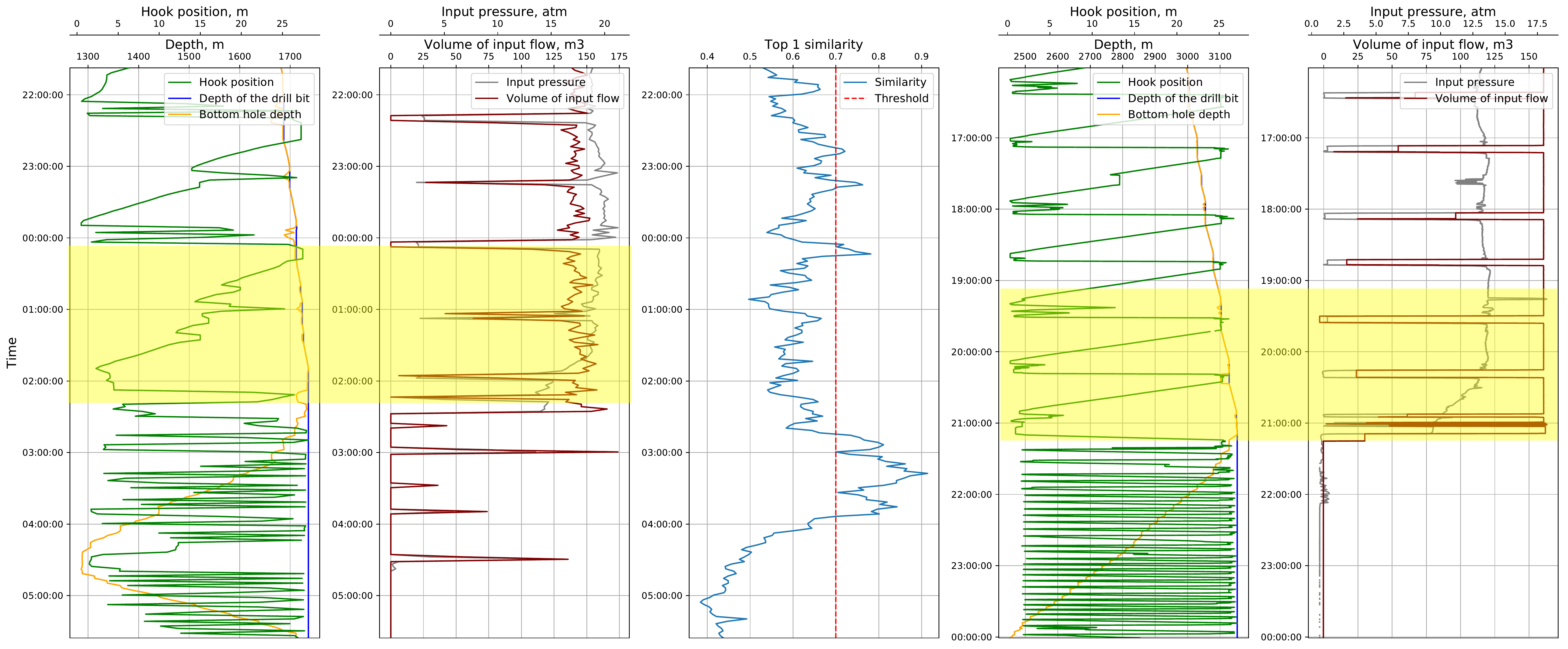}
    \caption{Analogues search model results in acion for hold-out well. Two plots on the right are MWD signals for a hold-out well, two plots on the left are the MWD for the analogue accident identified by the model. In the center there are similarity scores provided by the model: when similarity value exceeds the selected threshold the model alerts that the past two hours are similar to analogue measurements. For example, in the figure one of the similar areas for the hold-out case and analogue are highlighted in yellow colour. Both areas have similar signal trends, indicating a wash-out accident, which gives us an idea that the model correctly detected an accident and found a past analogue for it.}
    \label{fig:Example_analog_model}
\end{figure}

Figure \ref{fig:Example_analog_model} represented analogues search model results for one hold-out well, containing wash-out drilling accident. 

\section{}
\label{Appendix_1}
 
Table~\ref{tab:Total_tp_fp} provides more detailed results of the analogues search model testing.
 
\begin{table}[H]
\centering
\resizebox{\textwidth}{!}{%
\begin{tabular}{|l|l|l|l|l|l|l|l|l|l|l|l|l|l|l|}
    \hline
    Well & 
    \begin{tabular}[c]{@{}l@{}}True accident \\ type\end{tabular} & \begin{tabular}[c]{@{}l@{}}Total signal \\ duration \\ (days)\end{tabular} & \multicolumn{2}{l|}{Stuck} & \multicolumn{2}{l|}{Wash-outs} & \multicolumn{2}{l|}{Shale collars} & \multicolumn{2}{l|}{Mud loss} & \multicolumn{2}{l|}{Breaks} & \multicolumn{2}{l|}{Fluid shows} \\ \cline{4-15} 
     &  &  & $\mathcal{TP}$ & $\mathcal{FP}$ & $\mathcal{TP}$ & $\mathcal{FP}$ & $\mathcal{TP}$ & $\mathcal{FP}$ & $\mathcal{TP}$ & $\mathcal{FP}$ & $\mathcal{TP}$ & $\mathcal{FP}$ & $\mathcal{TP}$ & $\mathcal{FP}$ \\ \hline
    1 & Normal mode & 22 & 0 & 9 & 0 & 1 & 0 & 0 & 0 & 5 & 0 & 0 & 0 & 0 \\ \hline
    2 & Fluid show & 46 & 0 & 16 & 0 & 0 & 0 & 0 & 0 & 1 & 0 & 0 & 1 & 0 \\ \hline
    3 & Normal mode & 17 & 0 & 6 & 0 & 0 & 0 & 0 & 0 & 3 & 0 & 0 & 0 & 0 \\ \hline
    4 & Fluid show & 105 & 0 & 22 & 0 & 0 & 0 & 0 & 0 & 30 & 0 & 1 & 1 & 0 \\ \hline
    5 & Normal mode & 18 & 0 & 0 & 0 & 0 & 0 & 0 & 0 & 0 & 0 & 0 & 0 & 0 \\ \hline
    6 & Normal mode & 9 & 0 & 6 & 0 & 0 & 0 & 0 & 0 & 2 & 0 & 0 & 0 & 0 \\ \hline
    7 & Normal mode & 10 & 0 & 7 & 0 & 1 & 0 & 0 & 0 & 1 & 0 & 0 & 0 & 0 \\ \hline
    8 & Fluid show & 33 & 0 & 18 & 0 & 0 & 0 & 0 & 0 & 0 & 0 & 1 & 0 & 4 \\ \hline
    9 & Normal mode & 19 & 0 & 8 & 0 & 0 & 0 & 0 & 0 & 0 & 0 & 0 & 0 & 0 \\ \hline
    10 & Normal mode & 13 & 0 & 4 & 0 & 0 & 0 & 0 & 0 & 2 & 0 & 0 & 0 & 0 \\ \hline
    11 & Normal mode & 16 & 0 & 6 & 0 & 0 & 0 & 0 & 0 & 1 & 0 & 0 & 0 & 0 \\ \hline
    12 & Normal mode & 11 & 0 & 3 & 0 & 0 & 0 & 0 & 0 & 3 & 0 & 1 & 0 & 0 \\ \hline
    13 & \begin{tabular}[c]{@{}l@{}}Mud loss,\\  2 cases\end{tabular} & 31 & 0 & 17 & 0 & 0 & 0 & 0 & 1 & 3 & 0 & 0 & 0 & 0 \\ \hline
    14 & \begin{tabular}[c]{@{}l@{}}Wash-outs, \\ 2 cases\end{tabular} & 21 & 0 & 10 & 0 & 0 & 0 & 0 & 0 & 4 & 0 & 0 & 0 & 0 \\ \hline
    15 & Stuck & 35 & 1 & 28 & 0 & 0 & 0 & 0 & 0 & 2 & 0 & 0 & 0 & 0 \\ \hline
    16 & Stuck & 34 & 1 & 12 & 0 & 0 & 0 & 0 & 0 & 0 & 0 & 0 & 0 & 0 \\ \hline
    17 & Stuck & 27 & 1 & 13 & 0 & 0 & 0 & 0 & 0 & 1 & 0 & 0 & 0 & 0 \\ \hline
    18 & Stuck & 28 & 1 & 13 & 0 & 2 & 0 & 0 & 0 & 5 & 0 & 0 & 0 & 0 \\ \hline
    19 & Mud loss & 23 & 0 & 13 & 0 & 1 & 0 & 0 & 0 & 0 & 0 & 0 & 0 & 0 \\ \hline
    20 & Stuck & 34 & 1 & 12 & 0 & 2 & 0 & 0 & 0 & 3 & 0 & 0 & 0 & 0 \\ \hline
    21 & \begin{tabular}[c]{@{}l@{}}Wash-out, \\ Stuck 2 cases, \\ Mud loss 2 cases\end{tabular} & 31 & 1 & 7 & 0 & 0 & 0 & 0 & 0 & 0 & 0 & 0 & 0 & 0 \\ \hline
    22 & \begin{tabular}[c]{@{}l@{}}Wash-outs, \\ 2 cases\end{tabular} & 39 & 0 & 10 & 1 & 0 & 0 & 0 & 0 & 1 & 0 & 0 & 0 & 0 \\ \hline
    23 & Stuck & 31 & 0 & 11 & 0 & 0 & 0 & 0 & 0 & 0 & 0 & 0 & 0 & 0 \\ \hline
    24 & Normal mode & 30 & 0 & 13 & 0 & 0 & 0 & 0 & 0 & 0 & 0 & 0 & 0 & 0 \\ \hline
    25 & Shale collar & 18 & 0 & 8 & 0 & 0 & 0 & 0 & 0 & 6 & 0 & 0 & 0 & 0 \\ \hline
    26 & \begin{tabular}[c]{@{}l@{}}Shale collars, \\ 2 cases\end{tabular} & 15 & 0 & 3 & 0 & 0 & 0 & 0 & 0 & 0 & 0 & 0 & 0 & 0 \\ \hline
    27 & \begin{tabular}[c]{@{}l@{}}Wash-out, \\ Fluid show\end{tabular} & 27 & 0 & 20 & 0 & 0 & 0 & 0 & 0 & 1 & 0 & 0 & 0 & 0 \\ \hline
    28 & Shale collar & 41 & 0 & 15 & 0 & 0 & 0 & 0 & 0 & 3 & 0 & 0 & 0 & 0 \\ \hline
    29 & Fluid show & 57 & 0 & 16 & 0 & 1 & 0 & 0 & 0 & 8 & 0 & 0 & 1 & 0 \\ \hline
    30 & Fluid show & 101 & 0 & 43 & 0 & 0 & 0 & 1 & 0 & 1 & 0 & 0 & 1 & 0 \\ \hline
\end{tabular}}
\caption{Total number of correct and false alarms for each testing well and accident type for threshold $0.7$}
\label{tab:Total_tp_fp}
\end{table}

\end{appendices}

\end{document}